\documentclass[11pt]{article}

\usepackage[final]{acl}

\usepackage{times}
\usepackage{latexsym}
\usepackage[T1]{fontenc}
\usepackage[utf8]{inputenc}
\usepackage{microtype}
\usepackage{inconsolata}
\usepackage{graphicx}
\usepackage{booktabs}
\usepackage{tabularx}
\usepackage{multirow}
\usepackage{xcolor}
\usepackage{balance}
\usepackage{url}
\usepackage{hyperref}

\definecolor{llmblue}{HTML}{1B3A5C}
\definecolor{leftorange}{HTML}{C2410C}
\definecolor{consblue}{HTML}{1E40AF}
\definecolor{neutblue}{HTML}{D4A017}

\definecolor{gptgreen}{HTML}{D1F2DD}
\definecolor{deeppurple}{HTML}{E6DCF5}
\definecolor{llamablue}{HTML}{D6E8FB}
\definecolor{gpttext}{HTML}{0F7A40}
\definecolor{deeptext}{HTML}{6A2DAD}
\definecolor{llamatext}{HTML}{1A5FB4}
\newcommand{\gptbox}[1]{\colorbox{gptgreen}{#1}}
\newcommand{\dsbox}[1]{\colorbox{deeppurple}{#1}}
\newcommand{\llamabox}[1]{\colorbox{llamablue}{#1}}
\newcommand{\gpthl}[1]{\textcolor{gpttext}{#1}}
\newcommand{\dshl}[1]{\textcolor{deeptext}{#1}}
\newcommand{\llamahl}[1]{\textcolor{llamatext}{#1}}

\newcommand{\ourtool}{\mbox{\textsc{LLMpedia}}}
\newcommand{\site}{\url{https://llmpedia.net}}

\newcommand{\video}{\href{https://drive.google.com/file/d/1kBXB2cdUYye5Zo_j2Md4ckVo6j5-0Ni0/view?usp=sharing}{demonstration video}}

\newcommand{\live}[2]{\href{https://llmpedia.net/#2}{#1}}

\title{\ourtool{}: Browsing, Verifying, and Comparing\\
the Parametric Encyclopedic Knowledge of LLMs}

\author{
  \begin{tabular}{c}
    {\bf Muhammed Saeed} \qquad {\bf Simon Razniewski} \\
    ScaDS.AI Dresden/Leipzig \& TU Dresden, Germany \\
    \texttt{\{muhammed.saeed,simon.razniewski\}@tu-dresden.de}
  \end{tabular}
}

\begin{document}
\maketitle

% ═══════════════════════════════════════════════════════════════
% \begin{abstract}
% Flagship language models appear \emph{saturated} on benchmarks like
% MMLU \citep{hendrycks2021mmlu}, scoring above 90\% - yet benchmarks
% test only what the experimenter thought to ask, the
% \emph{availability bias} of fixed question sets. \ourtool{} makes
% this bias measurable, and browsable. We recursively materialized
% ${\sim}$1.3M encyclopedia articles from the parametric memory of
% three model families (GPT-5-mini, DeepSeek-V3.2, Llama-3.3-70B)
% without retrieval, then audited a stratified sample of atomic claims
% against Wikipedia and a curated web stack, coloring every claim
% \emph{supported}, \emph{refuted}, or \emph{insufficient}
% \citep{saeed2026llmpedia}. On a uniform random sample the verified
% true rate is 68.4\% - more than 21\,pp below MMLU - with 30.5\% of
% claims \emph{insufficient}: assertions no benchmark would ever probe
% and even the world's largest encyclopedia cannot adjudicate -
% long-tail knowledge or plausible hallucination, the evidence cannot
% tell - extending to free text the coverage gap GPTKB established for
% triples \citep{HuGPTKB2025}. The resulting live, fully open
% encyclopedia lets a visitor inspect this frontier one claim at a time
% through five one-click views - link-traversal exploration,
% claim-level factuality, cross-model and political-persona comparison,
% and a guided topic drill-down - each page, claim, and verdict at a
% stable URL. \ourtool{} is live at \site{}; see the \video{}.
% \end{abstract}
\begin{abstract}
Flagship language models appear \emph{saturated} on benchmarks like
MMLU \citep{hendrycks2021mmlu}, scoring above 90\% - yet benchmarks
test only what the experimenter thought to ask, the
\emph{availability bias} of fixed question sets. \ourtool{} makes
this bias measurable and browsable. We recursively materialized
${\sim}$1.3M articles from three model families' parametric memory
(GPT-5-mini, DeepSeek-V3.2, Llama-3.3-70B) without retrieval, then
audited a stratified sample of atomic claims against Wikipedia and a
curated web stack, coloring every claim \emph{supported},
\emph{refuted}, or \emph{insufficient} \citep{saeed2026llmpedia}. On
a uniform random sample the true rate is 68.4\% - more than 21\,pp
below MMLU - with 30.5\% of claims \emph{insufficient}: assertions no
benchmark probes and the world's largest encyclopedia cannot
adjudicate - long-tail knowledge or plausible hallucination, the
evidence cannot tell - extending to free text the coverage gap GPTKB
established for triples \citep{HuGPTKB2025}. The resulting live, open
encyclopedia lets visitors inspect this frontier one claim at a time
through five one-click views - link-traversal exploration,
claim-level factuality, cross-model and political-persona comparison,
and a guided topic drill-down - each page, claim, and verdict at a
stable URL. \ourtool{} is live at \site{}; see the \video{}.
\end{abstract}
% ═══════════════════════════════════════════════════════════════
\begin{figure}[t]
 \centering
 \includegraphics[width=\columnwidth]{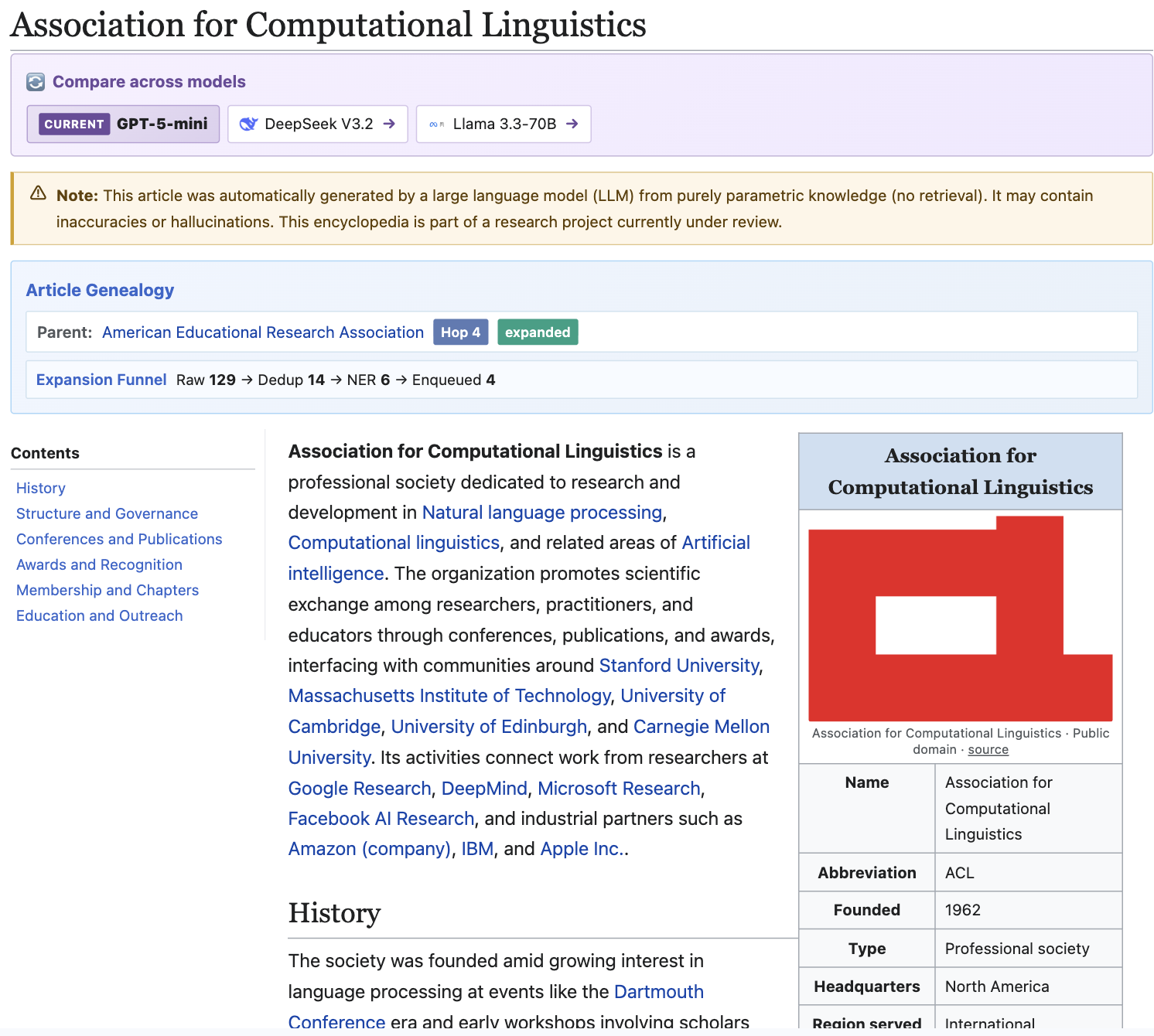}
 \caption{A materialized entity page for the
 \live{\emph{Association for Computational
 Linguistics}}{gpt-5-mini/Association_for_Computational_Linguistics.html}
 (ACL for short; GPT-5-mini): the \emph{Article Genealogy} panel
 (parent \emph{American Educational Research Association}, hop~4),
 the \emph{Expansion Funnel}, the infobox (founded 1962), and a lead
 paragraph of traversable wikilinks. Every blue term in this paper
 is a live link in the deployed site.}
 \label{fig:teaser}
\end{figure}

\begin{figure}[t]
 \centering
 \includegraphics[width=0.97\columnwidth,trim=0 0 0 0,clip]{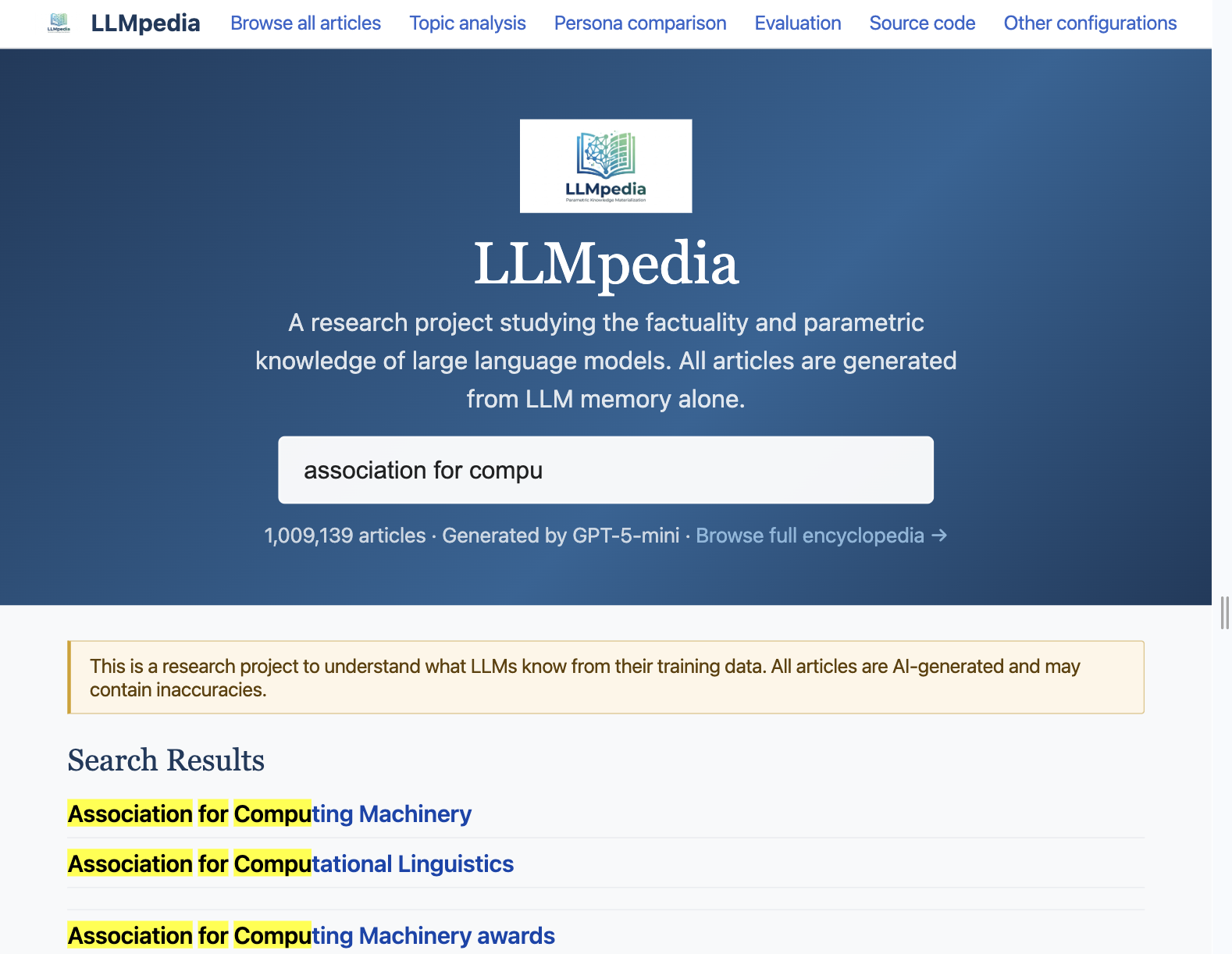}
 \caption{Searching \emph{association for compu} returns the entities
 GPT-5-mini surfaced \emph{on its own} - not a fixed index, but the
 model's own associations.}
 \label{fig:search}
\end{figure}
% ═══════════════════════════════════════════════════════════════
\section{Introduction}
\label{sec:intro}

Ask a large language model a multiple-choice question and you learn
one fact about one subject the experimenter chose. Yet LLMs encode
knowledge across an enormous space of subjects
\citep{petroni-etal-2019-language}, and standard instruments - MMLU
\citep{hendrycks2021mmlu}, TruthfulQA \citep{lin2022truthfulqa} -
probe only the slice someone thought to test. This is the
\emph{availability bias} of \citet{tversky1973availabilit}: a model
can look saturated on a fixed question set while vast regions of
weak, strong, or simply unverifiable knowledge go unmeasured - in
exactly the modality users read, fluent multi-paragraph prose, which
benchmarks never inspect.

\ourtool{} \citep{saeed2026llmpedia} takes the opposite route: let a
model write an entire encyclopedia from parametric memory alone - no
retrieval at generation time - then audit a sampled subset claim by
claim, turning latent knowledge into a \textbf{browsable, queryable,
fact-checkable} artifact. This submission is the interactive
demonstrator; the companion paper \citep{saeed2026llmpedia}
establishes methodology and findings. \emph{Here we show how a
visitor uses the live resource}, and why browsing it surfaces things
a leaderboard cannot.

\paragraph{The payoff is visible, not asserted.}
\ourtool{}'s central finding is that a model's dominant long-form
failure is \emph{silence}, not error: on a uniform random sample,
GPT-5-mini's claims are 68.4\% supported, only 1.2\% refuted, and
30.5\% \emph{unverifiable} \citep{saeed2026llmpedia}. A number is
easy to discount, so the demonstrator makes it tangible: open
\live{\emph{Annual Meeting of the
ACL}}{eval/Annual_Meeting_of_the_Association_for_Computational_Linguistics.html},
a hop-4 entity, and its claims appear colored \emph{supported}
(green), \emph{refuted} (orange), and \emph{insufficient} (grey),
each with evidence beside it. Browse toward the frontier and the mix
shifts before your eyes from mostly-green to mostly-grey -
availability bias, rendered.

\paragraph{Everything is one click away.}
Each entity has a stable, shareable URL
(\texttt{llmpedia.net/<model>/<Subject>.html}); the demonstrator is a
static, million-page site kept responsive by a client-side index. We
focus on five live use cases:
\textbf{(1)} link-based exploration of entities and relations
(\S\ref{sec:explore});
\textbf{(2)} claim-level factuality inspection on the audited subset
(\S\ref{sec:factuality});
\textbf{(3)} cross-model comparison of the same subject
(\S\ref{sec:crossmodel});
\textbf{(4)} persona comparison across editorial framings
(\S\ref{sec:persona}); and
\textbf{(5)} topic analysis, a guided topic/persona/model drill-down
(\S\ref{sec:topic}).

\noindent\textbf{Contributions.}
(i)~The first browsable, openly licensed (CC~BY~4.0) encyclopedia
materialized purely from LLM parametric memory at
${\sim}$1.3M-article scale, with stable per-entity URLs.
(ii)~An in-page audit view that makes the
\emph{supported}/\emph{refuted}/\emph{insufficient} distinction - and
thus availability bias - legible to a non-expert visitor.
(iii)~Side-by-side cross-model and cross-persona views that expose
knowledge divergence and framing effects a single-score leaderboard
hides. All prompts, articles, verdicts, and code are live at \site{}.

% ═══════════════════════════════════════════════════════════════
\section{\ourtool{} in One Minute}
\label{sec:method}

\paragraph{Article generation.}
From a single seed entity (\emph{Vannevar Bush}), the model writes a
full Wikitext article with \texttt{[[wikilinks]]}. New links pass
through three-stage sanitization - canonical normalization, LLM-based
encyclopedic filtering, and embedding-based deduplication - and
survivors are enqueued, each kept once under its first parent, in a
breadth-first expansion. The encyclopedia thus grows along the
model's \emph{own} associations: the structure a visitor browses is
the structure the model produced.

\paragraph{Claim verification.}
Per-claim verification is costly at million-article scale, so it runs
on a stratified audit sample. Each audited article is decomposed into
atomic claims; each claim is checked against evidence retrieved using
\emph{only the subject name}, so an article can never shape its own
evidence. Following the decompose--retrieve--verify paradigm
\citep{min-etal-2023-factscore, song-etal-2024-veriscore}, claims are
labeled \emph{supported}, \emph{refuted}, or \emph{insufficient} -
separating contradiction from silence is precisely what exposes
availability bias. The public audit covers 2{,}010 subjects and
20{,}092 claims; full methodology in \citet{saeed2026llmpedia}.

% ═══════════════════════════════════════════════════════════════
\section{Corpus and Numbers a Visitor Can Check}
\label{sec:construction}

We materialized three corpora: GPT-5-mini reaches ${\sim}$1M articles
in general-domain mode; DeepSeek-V3.2 and Llama-3.3-70B reach
${\sim}$120K each; all three additionally run in topic-focused mode
(\emph{Ancient Babylon}, \emph{US Civil Rights}, \emph{Dutch
Colonization}) for controlled comparison - ${\sim}$1.3M articles in
total (Table~\ref{tab:stats}). All runs use temperature-0 decoding
with fixed seeds, so every page on the site is a reproducible
artifact of its model.

Each entity enters the queue once and records the parent that first
surfaced it (\texttt{bfsParent}) and its hop distance
(\texttt{bfsLayer}); these power the genealogy panel
(\S\ref{sec:explore}). Audited claims are checked against two
evidence tiers: Wikipedia, and a curated web tier of 133
quality-scored domains (plus them we favor endings like \texttt{.gov} and
\texttt{.edu}; \citealp{saeed2026llmpedia}) that reaches the
${\sim}$43\% of subjects Wikipedia does not cover. On a
uniform-random sample, GPT-5-mini reaches a 68.4\% true rate, with
unverifiability (30.5\%), not falsehood (1.2\%), the dominant
non-true outcome - a result the audit view lets a visitor re-derive
by hand on any sampled page.

\begin{table}[h!]
\centering
\small
\setlength{\tabcolsep}{4pt}
\renewcommand{\arraystretch}{1.12}
\begin{tabularx}{\columnwidth}{@{}Xl@{}}
\toprule
\textbf{Articles} & ${\sim}$1.3M (3 model families) \\
\textbf{GPT-5-mini} & ${\sim}$1M articles (general domain) \\
\textbf{Open-weight} & ${\sim}$120K each (DeepSeek, Llama) \\
\textbf{Audited subset} & 2{,}010 subjects / 20{,}092 claims \\
\textbf{Verdicts} & supported / refuted / insufficient \\
\textbf{Evidence} & Wikipedia + curated domains \\
\textbf{Wikipedia coverage} & 56.7\% of surfaced subjects \\
\textbf{True rate (random)} & 68.4\% (1.2\% false, 30.5\% unv.) \\
\textbf{True rate (frontier)} & 57.6\% (web evidence) \\
\textbf{3-model overlap} & 7.3\% of subjects \\
\textbf{License} & CC BY 4.0 (all artifacts) \\
\bottomrule
\end{tabularx}
\caption{\ourtool{} at a glance \citep{saeed2026llmpedia}. Verdicts
are computed on a stratified audit sample, not the full corpus.}
\label{tab:stats}
\end{table}

% ═══════════════════════════════════════════════════════════════
\section{Web Provision and Access}
\label{sec:web}

\ourtool{} is hosted at \site{} as a static site: ${\sim}$1M articles
render as plain HTML with no server-side database, and \emph{search}
runs client-side over a compact in-browser inverted index, so the
site stays responsive at million-article scale. Each entity has a
stable URL \texttt{llmpedia.net/<model>/<Subject>.html} for deep
linking and citation. \emph{Provenance} is on every page: the
\emph{Article Genealogy} panel shows \texttt{bfsParent} and
\texttt{bfsLayer}, and the \emph{Expansion Funnel} reports how
outbound links flowed through the pipeline
(Raw~$\rightarrow$~Dedup~$\rightarrow$~NER~$\rightarrow$~Enqueued),
making construction itself inspectable. \emph{Analysis} ships as a
cross-model strip on every article and a persona-comparison page,
plus a CC~BY~4.0 dump and source.

% ═══════════════════════════════════════════════════════════════
\section{Demonstration Experience}
\label{sec:demo}

The five views below each isolate one axis; the topic view
(\S\ref{sec:topic}) ties them together. Throughout, blue terms are
live links - a reader can click straight into the demonstrator.

% ───────────────────────────────────────────────────────────────
\subsection{Link-Based Knowledge Exploration}
\label{sec:explore}

A visitor reaches an entity three ways: seed links on the start page,
the search field, or a direct URL. Typing \emph{association for
compu} (Figure~\ref{fig:search}) returns the entities the model
surfaced on its own; clicking the \emph{Association for Computational
Linguistics} (ACL) opens its page (Figure~\ref{fig:teaser}) - the
society GPT-5-mini describes as founded in 1962, with its own outline
spanning history, governance, conferences, and awards. From there a
visitor walks the graph \emph{outward} by clicking any wikilink,
e.g.\ from
\live{\emph{ACL}}{gpt-5-mini/Association_for_Computational_Linguistics.html}
to the \live{\emph{ACL Anthology}}{gpt-5-mini/ACL_Anthology.html} or
\live{\emph{EMNLP}}{gpt-5-mini/EMNLP.html}.

Because each entity was discovered as the child of exactly one
parent, the genealogy panel also lets a visitor move \emph{up} the
layers via \texttt{bfsParent}, tracing provenance back toward the
seed. This two-axis navigation - outward by association, upward by
provenance - is unique to a materialized encyclopedia. The panel
further exposes which of two lifecycle states a page is in. An
\textbf{expanded} page had its outbound wikilinks pushed through the
sanitization funnel, so the panel reports the full
Raw~$\rightarrow$~Dedup~$\rightarrow$~NER~$\rightarrow$~Enqueued
counts and the page's children are themselves browsable - e.g.\ the
\live{\emph{GPT-5-mini}}{gpt-5-mini/Association_for_Computational_Linguistics.html}
and \live{\emph{Llama}}{llama/Association_for_Computational_Linguistics.html}
ACL pages of Figure~\ref{fig:compare}, whose funnels read
129$\,\to\,$14$\,\to\,$6$\,\to\,$4 and
84$\,\to\,$16$\,\to\,$14$\,\to\,$11. A \textbf{terminal} page sits at
the crawl frontier: its article was fully generated, but the run
ended before its links were processed, so no funnel is shown and it
has no children of its own - e.g.\
\live{\emph{DeepSeek's}}{deepseek/Association_for_Computational_Linguistics.html}
ACL page, accepted at hop~4 with its outbound links never
NER-processed. Terminal thus marks where the \emph{expansion}
stopped, not where the model's knowledge ends - the same page would
grow children in a longer run.

% ───────────────────────────────────────────────────────────────
\subsection{Inspecting Factuality at the Claim Level}
\label{sec:factuality}

\begin{figure}[t]
 \centering
 \includegraphics[width=\columnwidth,trim=0 10 15 10,clip]{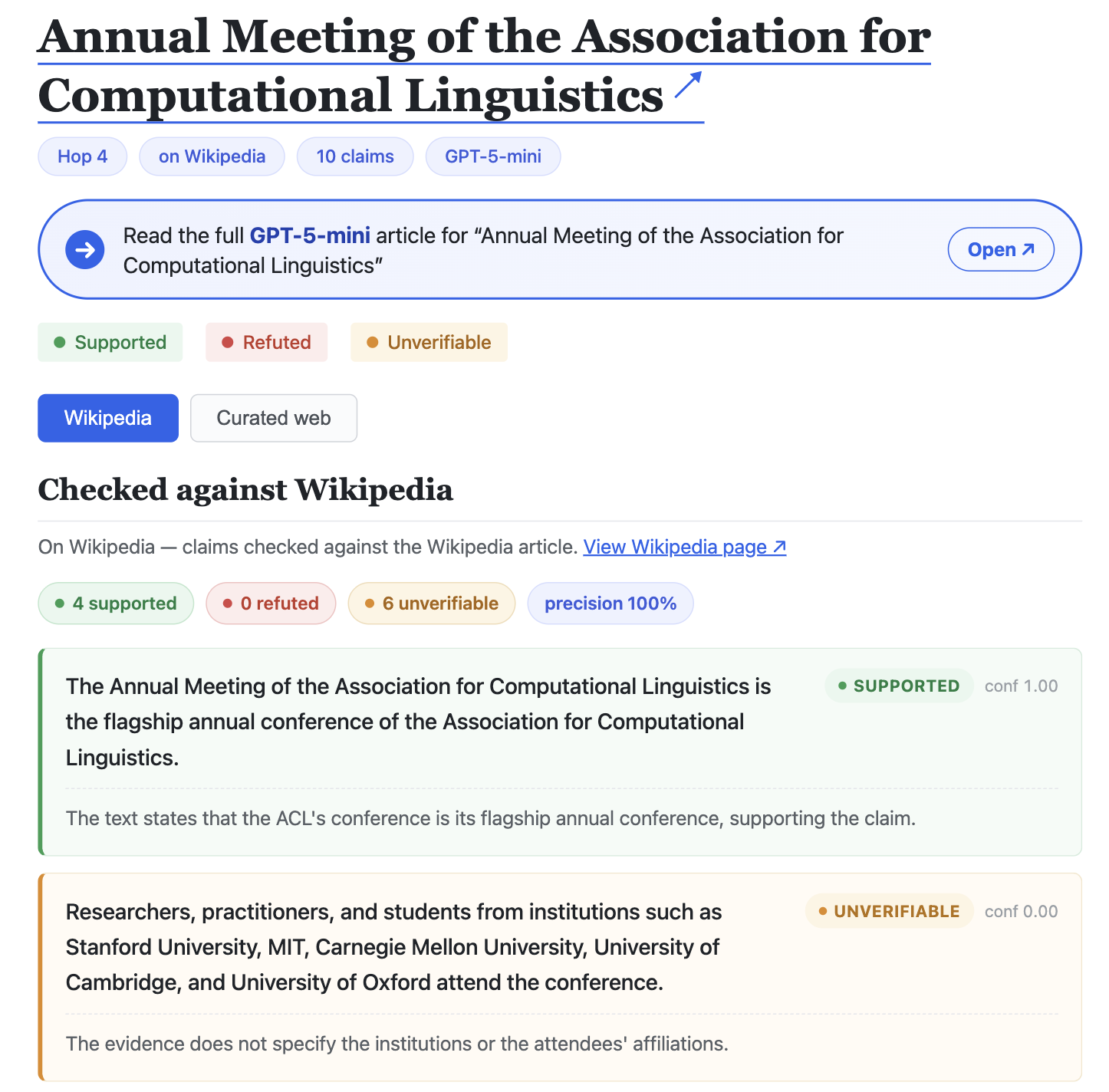}
 \caption{Claim-level evaluation of
 \live{\emph{Annual Meeting of the ACL}}{eval/Annual_Meeting_of_the_Association_for_Computational_Linguistics.html}
 (hop-4, GPT-5-mini): 10 atomic claims, each with a verdict
 (\textcolor{gpttext}{\emph{supported}}~/
 \textcolor{leftorange}{\emph{refuted}}~/ \emph{unverifiable}) and
 inline evidence. On Wikipedia the mix is 4/0/6 (precision~100\%);
 toggling to curated web lifts five unverifiable claims to supported.}
 \label{fig:eval}
\end{figure}

This view turns a statistic into an experience. Audited pages are
reachable from the site's \emph{Evaluation} tab; for every article in
the audit sample, each atomic claim carries a verdict
(Figure~\ref{fig:eval}), so a visitor sees not just \emph{whether} a
claim holds but \emph{why} it is hard to check. The audit spans
2{,}010 subjects and 20{,}092 claims (1{,}277 subjects verified on
Wikipedia, 733 on the curated-web frontier). Table~\ref{tab:hop}
previews the gradient a visitor can browse: as BFS depth grows, true
rate falls (94.0\%~$\rightarrow$~56.0\%) while the false rate stays
under 2\%; what rises is the \emph{unverifiable} rate, tracking the
collapse in Wikipedia coverage. The model is not asserting more
falsehoods at depth - external evidence simply cannot reach the
long-tail knowledge it encodes.

\begin{table}[t]
\centering
\small
\setlength{\tabcolsep}{4pt}
\renewcommand{\arraystretch}{1.05}
\begin{tabular}{lrrccc}
\toprule
\textbf{Bucket} & \textbf{Cov\%} & \textbf{Prec} & \textbf{True}
  & \textbf{False} & \textbf{Unv} \\
\midrule
hop~1    & 100  & 97.9 & \textbf{94.0} & 2.0 & 4.0 \\
hop~3    & 84   & 98.7 & \textbf{80.3} & 0.6 & 19.1 \\
hop~6    & 51   & 96.5 & \textbf{56.0} & 1.0 & 43.0 \\
\midrule
random   & 56.7 & 97.1 & 68.4 & 1.2 & 30.5 \\
frontier & 71.8 & 98.3 & 57.6 & 0.6 & 41.8 \\
\bottomrule
\end{tabular}
\caption{GPT-5-mini factuality by BFS depth, plus random and frontier
rows (\emph{frontier} = absent from Wikipedia, web evidence), on the
audited subset; all rates \% \citep{saeed2026llmpedia}.}
\label{tab:hop}
\end{table}

% ───────────────────────────────────────────────────────────────
\subsection{Cross-Model Analysis}
\label{sec:crossmodel}

\begin{figure*}[t]
  \centering
  \begin{minipage}[t]{0.325\textwidth}
    \centering
    {\small\textbf{\live{\gptbox{GPT-5-mini}}{gpt-5-mini/Association_for_Computational_Linguistics.html}}}\\[2pt]
    \includegraphics[width=\linewidth,trim=0 8 0 0,clip]{figures/acl_gpt_5_mini.png}
  \end{minipage}
  \hfill
  \begin{minipage}[t]{0.325\textwidth}
    \centering
    {\small\textbf{\live{\dsbox{DeepSeek V3.2}}{deepseek/Association_for_Computational_Linguistics.html}}}\\[2pt]
    \includegraphics[width=\linewidth,trim=0 8 0 0,clip]{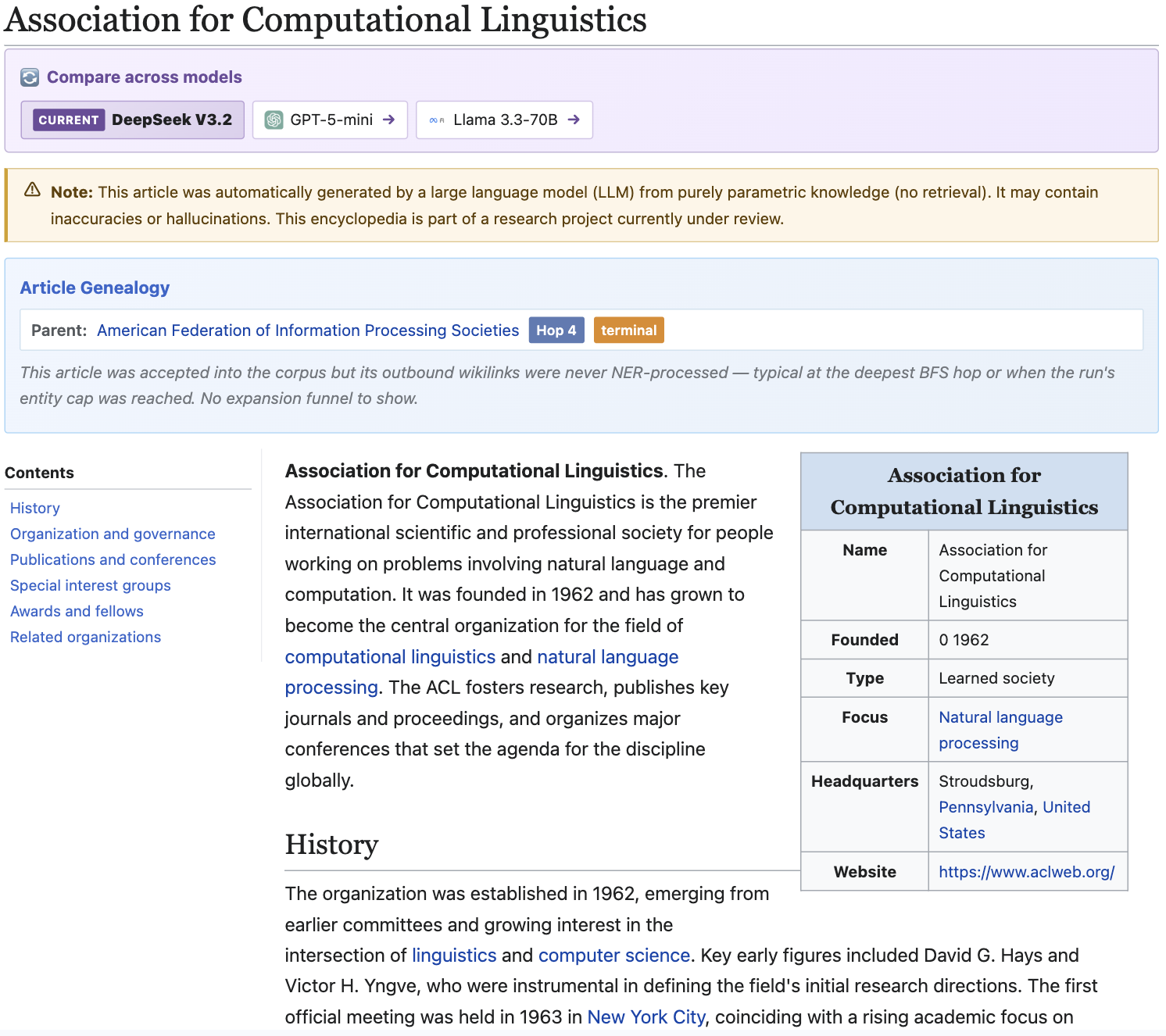}
  \end{minipage}
  \hfill
  \begin{minipage}[t]{0.325\textwidth}
    \centering
    {\small\textbf{\live{\llamabox{Llama 3.3-70B}}{llama/Association_for_Computational_Linguistics.html}}}\\[2pt]
    \includegraphics[width=\linewidth,trim=0 8 0 0,clip]{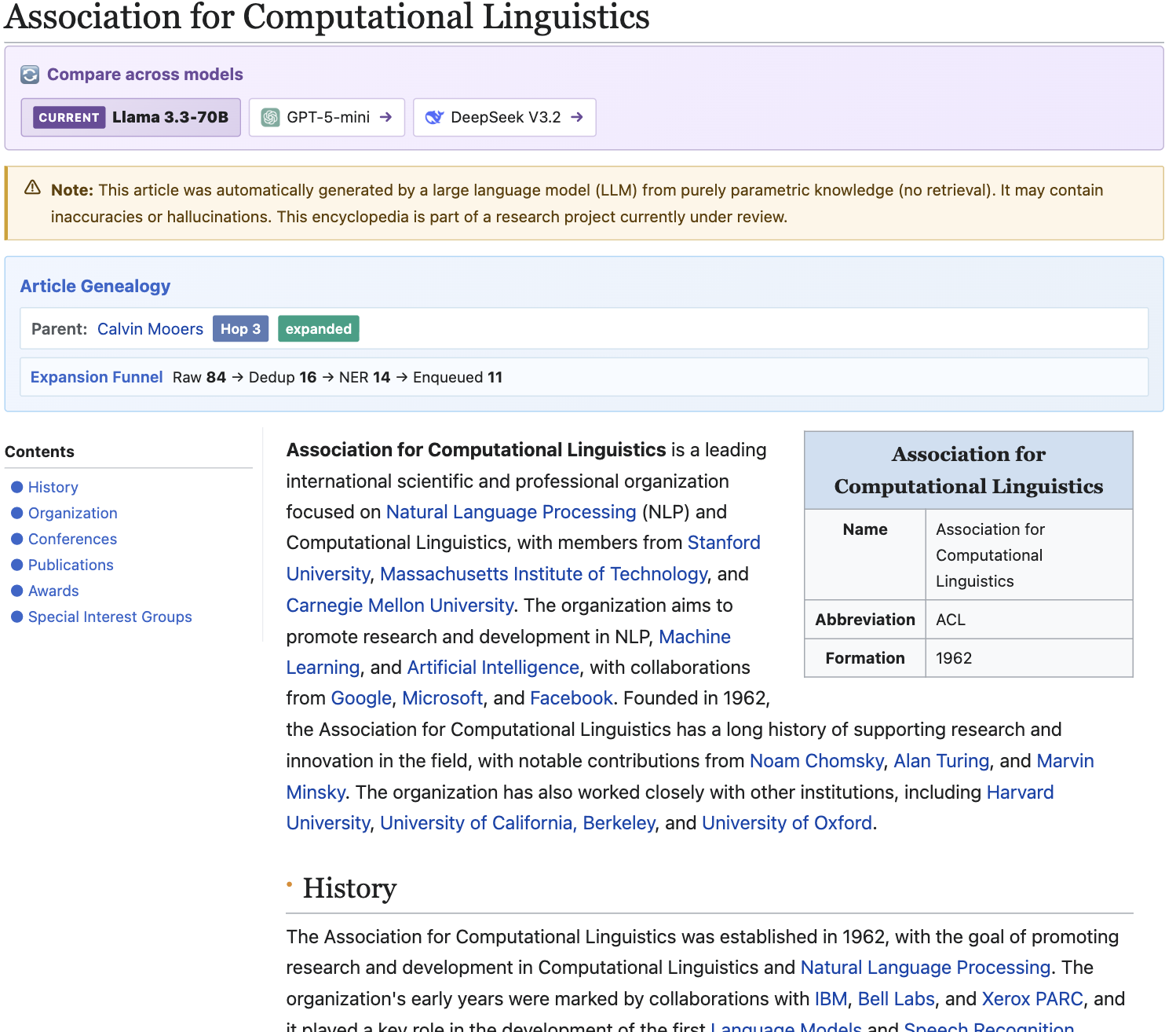}
  \end{minipage}
  \caption{Cross-model view of the
  \live{\emph{Association for Computational
  Linguistics}}{gpt-5-mini/Association_for_Computational_Linguistics.html}:
  different outlines, different infobox schemas, and entirely
  different BFS paths - GPT-5-mini via \live{\emph{American
  Educational Research
  Association}}{gpt-5-mini/American_Educational_Research_Association.html}
  (hop~4), DeepSeek via \live{\emph{American Federation of
  Information Processing
  Societies}}{deepseek/American_Federation_of_Information_Processing_Societies.html}
  (hop~4), Llama via \live{\emph{Calvin
  Mooers}}{llama/Calvin_Mooers.html} (hop~3). The \emph{Compare
  across models} strip atop every article switches between these
  views in one click.}
  \label{fig:compare}
\end{figure*}

Beyond browsing a single model, it is natural to ask how the same
subject changes when the model does. \ourtool{} materializes the
three families at deliberately different scales: GPT-5-mini, accessed
via API, was expanded to ${\sim}$1M articles, while the open-weight
DeepSeek-V3.2 and Llama-3.3-70B were run on on-premise GPUs and reach
${\sim}$120K articles each - so the cross-model views compare the
models on the subjects they share \citep{saeed2026llmpedia}. The
\emph{Compare across models} strip on every page re-renders the
current subject as written by another model. Figure~\ref{fig:compare}
shows the ACL by \gpthl{GPT-5-mini}, \dshl{DeepSeek-V3.2}, and
\llamahl{Llama-3.3-70B}; the three differ at every level.
\textbf{Infoboxes:} all agree the ACL was \emph{founded in 1962}, but
expose different schemas - \gpthl{GPT-5-mini} lists
\gpthl{Abbreviation, Type, Headquarters, Region served}, while the
open-weight models emit leaner or differently keyed boxes.
\textbf{Outlines:} each model proposes its own section structure,
consistent with corpus-level section-count differences
\citep{saeed2026llmpedia}. \textbf{Reliability:} the stable ranking
GPT-5-mini\,$>$\,DeepSeek\,$>$\,Llama holds - precision persists from
the shared core to each model's long tail (max drop 1.1\,pp) but true
rate falls steeply (88.6$\to$78.6, 85.8$\to$77.6, 79.1$\to$64.5) as
Wikipedia coverage thins. \textbf{Provenance:} the genealogy panels show the models even
\emph{reached} ACL differently - and expose their lifecycle badges:
the GPT-5-mini and Llama pages are \textbf{expanded} (each with its
own funnel), while DeepSeek's is \textbf{terminal}, its outbound
links never processed before the run stopped (\S\ref{sec:explore}) -
so even the crawl frontier differs per model on the same subject. At
scale this compounds: from the shared \emph{Vannevar Bush} seed, only 7.3\% of
subjects appear in all three corpora (entity Jaccard 0.17--0.22) -
models foreground \emph{entirely different entities}, a divergence
invisible to a leaderboard but immediate side by side
\citep{saeed2026llmpedia}.

% ───────────────────────────────────────────────────────────────
\subsection{Persona Comparison}
\label{sec:persona}

\begin{figure*}[h!]
  \centering
  \includegraphics[width=\textwidth,trim=0 100 0 0,clip]{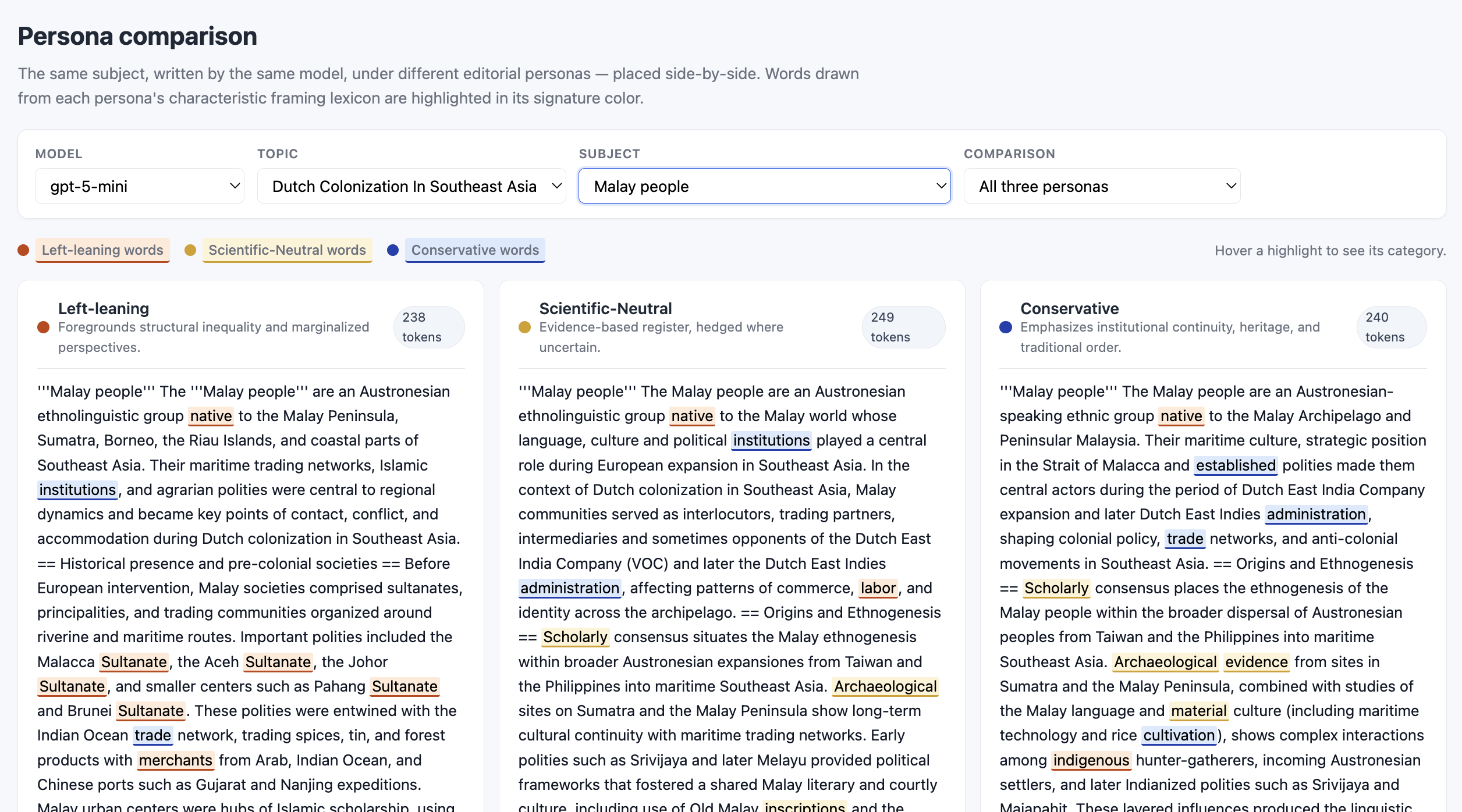}
  \caption{\live{\emph{Persona comparison}}{persona_comparison.html}
  for \emph{Malay people} under \emph{Dutch Colonization} (GPT-5-mini):
  the same subject under three personas, side by side. A deterministic
  classifier over a 24-dimensional framing lexicon
  \citep{saeed2026llmpedia} highlights matches in each persona's color
  ({\color{leftorange}left-leaning}, {\color{neutblue}scientific-neutral},
  {\color{consblue}conservative}) and tallies per-column hits: 13
  left-lexicon hits vs.\ 2 in the neutral column, at unchanged factual
  precision - only the framing shifts.}
  \label{fig:persona}
\end{figure*}

\begin{figure*}[h!]
  \centering
  \begin{minipage}[t]{0.40\textwidth}
    \centering
    {\scriptsize\textbf{1. Pick a topic}}\\[1pt]
    \includegraphics[width=\linewidth,trim=0 6 0 6,clip]{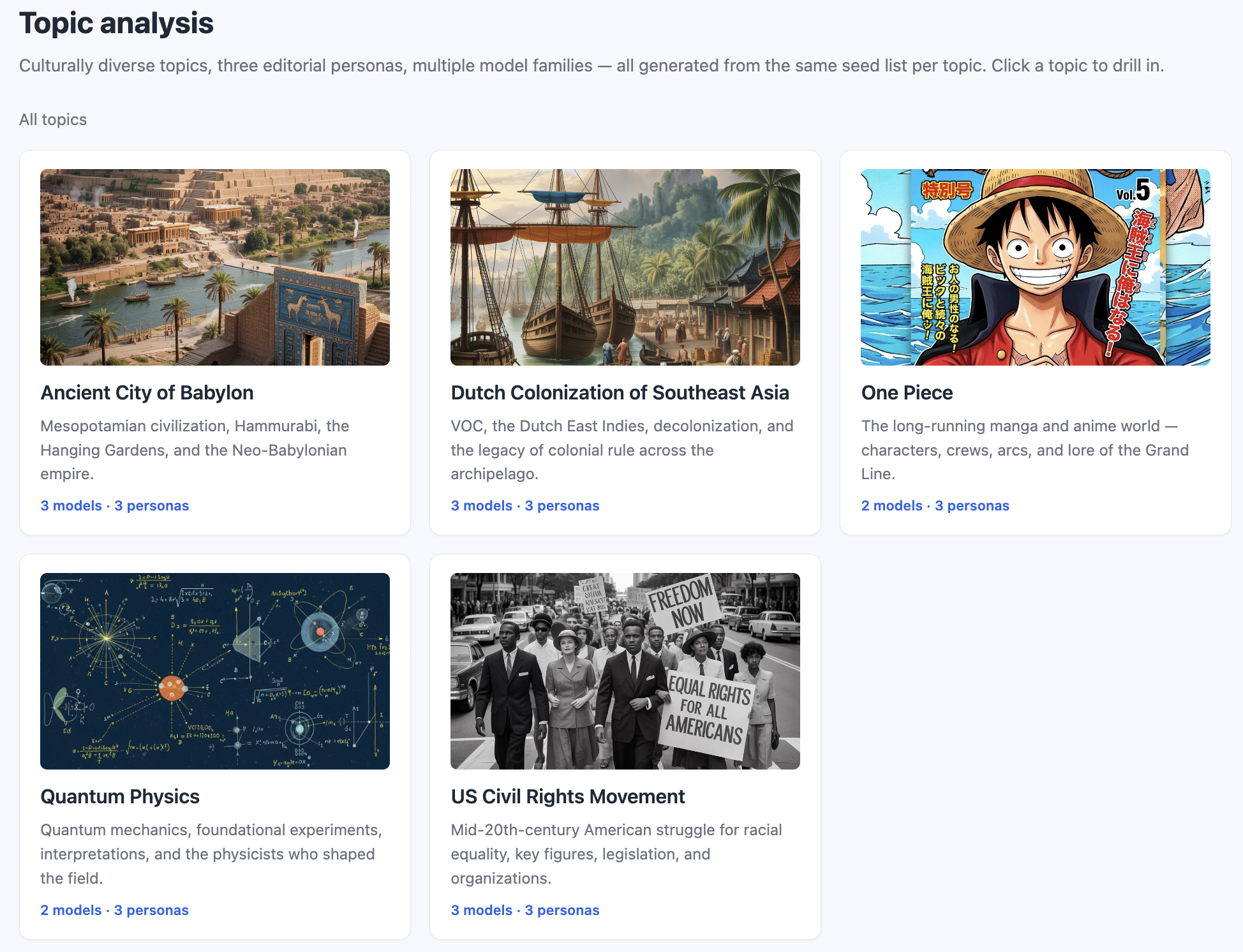}\\[4pt]
    {\scriptsize\textbf{2. Pick a persona}}\\[1pt]
    \includegraphics[width=\linewidth,trim=0 6 0 6,clip]{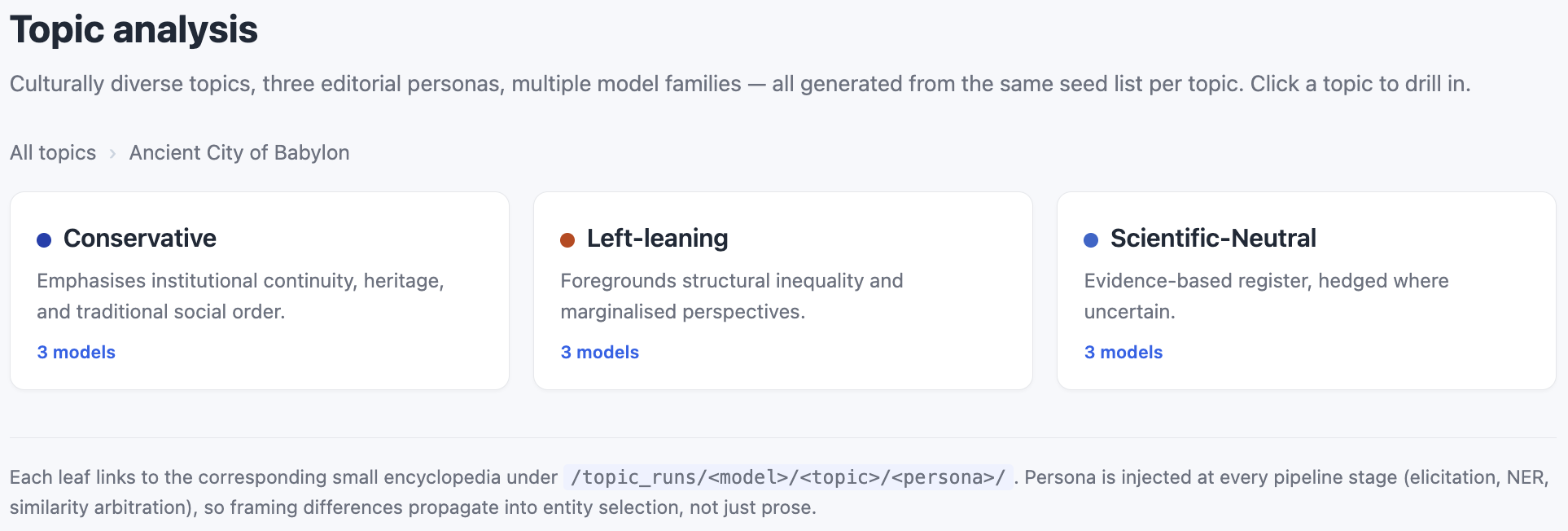}\\[4pt]
    {\scriptsize\textbf{3. Pick a model}}\\[1pt]
    \includegraphics[width=\linewidth,trim=0 6 0 6,clip]{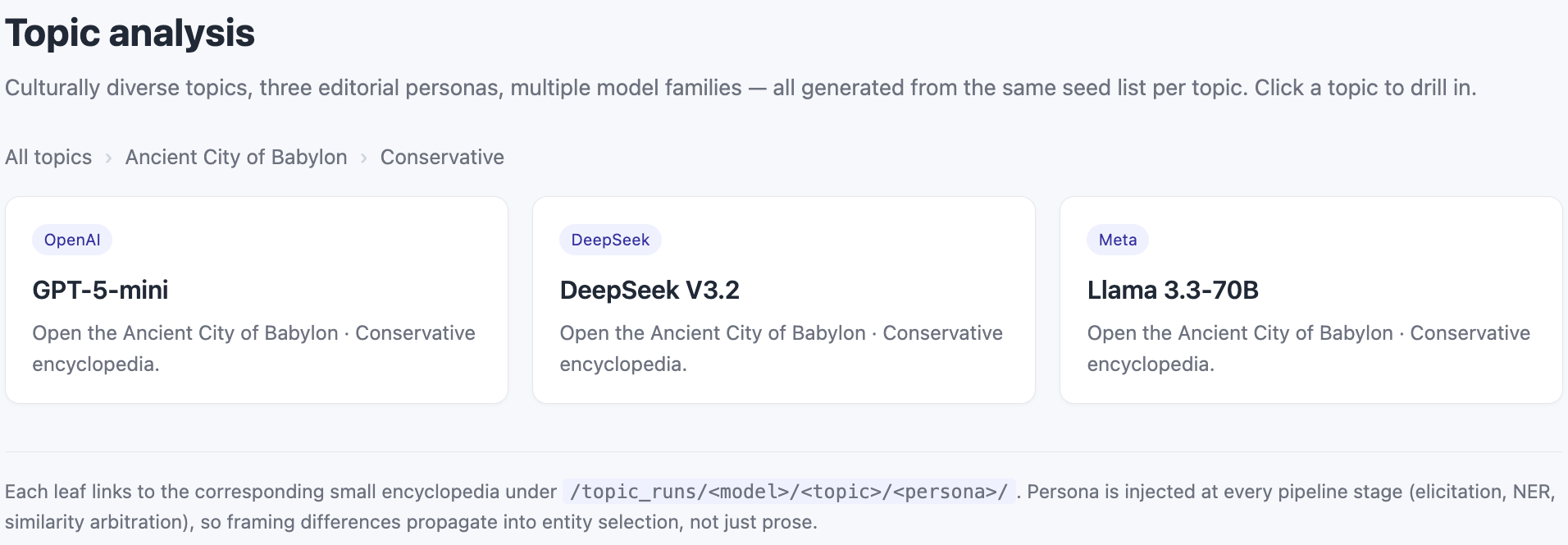}
  \end{minipage}
  \hfill
  \begin{minipage}[t]{0.55\textwidth}
    \centering
    {\scriptsize\textbf{4. Read the article}}\\[1pt]
    \includegraphics[width=\linewidth,trim=0 0 20 0,clip]{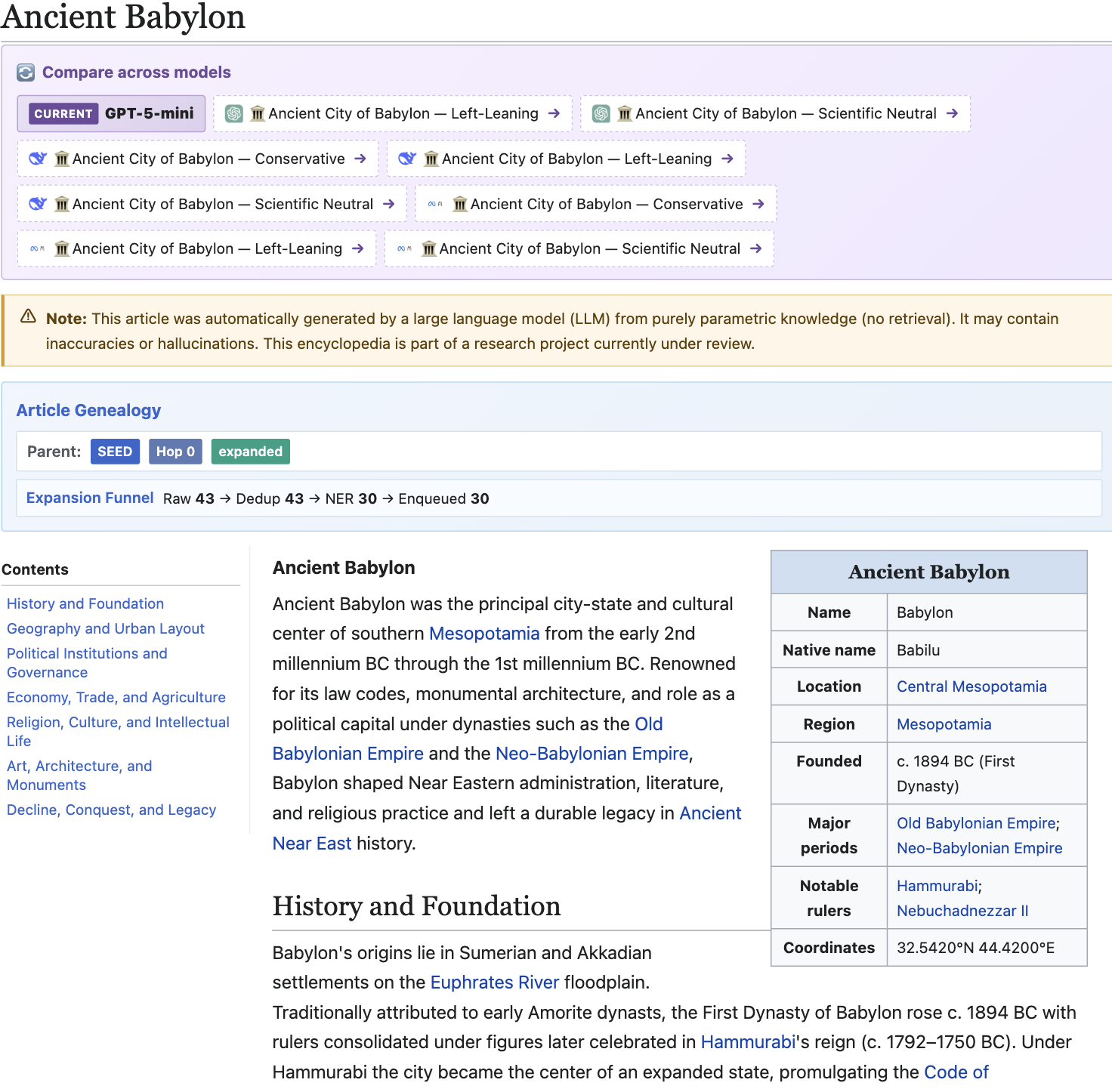}
  \end{minipage}
  \caption{The \live{\emph{Topic analysis}}{topic_analysis.html} view:
  pick a topic (1), a persona (2), a model (3), and land on the
  corresponding small encyclopedia (4). Every leaf is a
  \texttt{/topic\_runs/<model>/<topic>/<persona>/} run with the seed
  list held fixed, so a visitor can isolate the effect of each axis.
  The example walks \live{\emph{Ancient City of
  Babylon}}{topic_runs/gpt-5-mini/ancient_babylon/conservative/Ancient_Babylon.html}
  $\rightarrow$ \emph{Conservative} $\rightarrow$ \emph{GPT-5-mini}.}
  \label{fig:topic}
\end{figure*}

Whether an LLM-written encyclopedia carries an ideological slant is
no hypothetical: Grokipedia has drawn exactly this criticism
\citep{yasseri2025similar}. \ourtool{}
therefore treats framing as an explicit, \emph{deterministically
measured} variable rather than an accusation. The \emph{Persona
comparison} page contrasts the same subject, same
model, under three editorial personas -
{\color{leftorange}\textbf{left-leaning}} (structural inequality,
marginalized perspectives), {\color{neutblue}\textbf{scientific-neutral}}
(evidence-based, hedged), and {\color{consblue}\textbf{conservative}}
(institutional continuity, heritage) - injected at every pipeline
stage. Each article is scored by a word-list classifier over a
24-dimensional framing lexicon (hits per 1{,}000 tokens;
\citealp{saeed2026llmpedia});
matches are colored in their persona's hue with per-column counts,
and dropdowns pivot across model, topic, subject, and persona pair.
Only subjects present under all three personas are listed, so the
columns stay directly comparable. Personas are injected in the
topic-focused runs (\S\ref{sec:topic}) on two topics chosen because
framing effects are likely - \emph{Dutch Colonization in Southeast
Asia}, where personas can affect whether colonizer or colonized
perspectives are foregrounded, and the \emph{US Civil Rights}
movement - with \emph{Ancient City of Babylon} as the control:
ancient history, where writing should stay
neutral.\footnote{Two further low-contestedness controls,
\emph{Quantum Physics} and the \emph{One Piece} anime, are available
for GPT-5-mini and Llama only \citep{saeed2026llmpedia}.}

Figure~\ref{fig:persona} shows \emph{Malay people}. All three assert
the same facts - Austronesian origins, the Malacca and Aceh
sultanates, VOC contact - yet the count makes framing concrete: 13
left-lexicon hits (\emph{kinship}, \emph{customary law},
\emph{rulers}) against 2 in the neutral column, whose register
foregrounds \emph{scholarly} and \emph{archaeological}, while
conservative leans on \emph{trade}, \emph{religion}, \emph{material}.
This is the visible face of a corpus-level result: paired Wilcoxon
tests (Bonferroni-corrected over 648 comparisons) yield 37
significant persona effects in expected directions - e.g.\ on Dutch
Colonization, left-leaning uses colonized-side vocabulary
(\emph{exploitation}, \emph{plunder}, \emph{dispossession}) at
$+5.6$ hits/1{,}000 tokens over conservative, which flips to
development framing - yet precision is essentially unchanged
($\leq 3.6$\,pp within a cell), and on the neutral and control topics
the contested axes largely collapse (6 significant shifts vs.\ 37).
Persona changes \emph{what} an article emphasizes, not \emph{how
often it is right} \citep{saeed2026llmpedia}.

% ───────────────────────────────────────────────────────────────
\subsection{Topic Analysis: A Guided Drill-Down}
\label{sec:topic}

The \emph{Topic analysis} page connects the four views into a single
narrative: the three primary topics and three personas of
\S\ref{sec:persona}, crossed with the three model families, all grown
from the same per-topic seed list. A visitor drills down in three
clicks (Figure~\ref{fig:topic}): topic, then persona, then model.

The design is what makes the comparison controlled: with the seed
list held fixed per leaf, any downstream difference is attributable
to the model or persona, not a different starting point. Because
persona is injected at \emph{every} stage, framing propagates into
\emph{entity selection}, not just prose. The landing article carries
the same genealogy, funnel, and cross-model controls as any page,
handing the visitor back to the other views on a controlled slice.
Full-page browsing
also surfaces generation pathologies benchmarks never see: while
\gpthl{\live{GPT-5-mini's \emph{Ancient City of
Babylon}}{topic_runs/gpt-5-mini/ancient_babylon/conservative/Ancient_Babylon.html}}
reads fluently, the
\dshl{\live{DeepSeek run}{topic_runs/deepseek/ancient_babylon/conservative/Ancient_Babylon.html}}
collapses into a degenerate repetition loop - visible immediately in
the side-by-side comparison.

% ═══════════════════════════════════════════════════════════════
\section{Related Work}
\label{sec:related}

Factual LLM knowledge is studied mostly via sample-based probes such
as LAMA \citep{petroni-etal-2019-language} and benchmarks like MMLU
\citep{hendrycks2021mmlu}, which measure only what the experimenter
thought to ask. Materialization instead surfaces knowledge on the
model's terms: \citet{cohen2023crawling} introduced recursive
elicitation, and GPTKB \citep{HuGPTKB2025} scaled it to 100M triples.
\ourtool{} \citep{saeed2026llmpedia} extends materialization from
triples to discourse-level \emph{articles}, the modality users
actually read. Verification follows the decompose--retrieve--verify
line \citep{min-etal-2023-factscore, wei2024safe,
song-etal-2024-veriscore}. Among generated encyclopedias, STORM
\citep{shao-etal-2024-assisting} is retrieval-grounded, whereas
\ourtool{} is purely parametric; Grokipedia \citep{grokipedia}
operates at scale but discloses no methodology
\citep{yasseri2025similar}.

% % ═══════════════════════════════════════════════════════════════
% \section{Conclusion}
% \label{sec:conclusion}
%  We presented \ourtool{}, a ${\sim}$1.3M-article encyclopedia from three model families. Claim-level audits show that models fail mainly through unverifiable claims---long-tail knowledge or plausible hallucination - rather than outright error. All artifacts are available at \site{}.

\section{Conclusion}
\label{sec:conclusion}
 
We presented \ourtool{}, a ${\sim}$1.3M-article  encyclopedia materialized from
the parametric memory of three model families and audited
claim-by-claim on a sampled subset \citep{saeed2026llmpedia}. It
shows what benchmarks cannot: a model's dominant failure is silence,
not error. All artifacts are at \site{}.

% ═══════════════════════════════════════════════════════════════
\section{Limitations}

\paragraph{Single-pass generation.} Each article is one temperature-0
generation under a fixed prompt, so it samples what a model surfaces,
not the full extent of its parametric knowledge; an omitted fact is
not evidence the model lacks it. Repeated sampling
\citep{wang2022self} or single-claim elicitation
\citep{petroni-etal-2019-language, sun2024headtail} would give a
fuller picture at much higher cost.

\paragraph{Sampled verification and judge dependence.} Verdicts cover
a stratified audit of 2{,}010 subjects and 20{,}092 claims, not all
${\sim}$1.3M articles; unaudited pages are browsable but unverified.
The LLM judge (\texttt{gpt-4.1-nano}) is validated against human
annotations and FActScore \citep{saeed2026llmpedia}, but residual
error remains, and \emph{insufficient} does not imply falsehood; LLM
judges are nonetheless common practice in long-form factuality
evaluation \citep{min-etal-2023-factscore, song-etal-2024-veriscore,
rajendhran-etal-2025-verifastscore}.

\paragraph{Temporal and evidence limits.} The corpus is a
January--March 2026 snapshot; as sources, models, and Wikipedia
evolve, verdicts may shift. Under strict source-quality criteria,
28.2\% of frontier subjects return no usable web evidence, biasing
Tier~2 coverage toward well-documented entities.

\paragraph{Unequal model scale.} GPT-5-mini reaches ${\sim}$1M
articles versus ${\sim}$120K for each open-weight model, so the
deep-hop and long-tail analyses are GPT-5-mini specific.

\paragraph{Hallucinated entities.} Recursive expansion can surface
subjects whose facts, relations, or existence are hallucinated or
conflated; a page's presence is not evidence its subject exists or
that its content is accurate, as disclosed prominently on \site{}.

\paragraph{Entity resolution.} Sense arbitration keeps
differently-worded homonyms apart (the
\live{\emph{ACL}}{gpt-5-mini/ACL.html} of the knee vs.\ the
spelled-out \emph{Association for Computational Linguistics}), but
distinct entities sharing one \emph{unqualified} surface form (e.g.\
different places named \emph{Dresden}) may merge before arbitration
runs; resolving this needs context-aware, sense-specific linking.

\paragraph{Scope of the metric.} We measure whether individual
propositions are evidence-supported - not coherence, neutrality,
completeness, salience, or omission; a high-precision article can
still be incomplete or misleadingly organized. \ourtool{} studies
surfaced knowledge; it does not certify article quality.

% ═══════════════════════════════════════════════════════════════
\section{Ethics Statement}

\ourtool{} \citep{saeed2026llmpedia} probes what language models can
surface from parametric memory alone, locating where model knowledge
is weak, unverifiable, or wrong - and showing that benchmark
saturation does not imply broad or reliable long-tail knowledge.

\ourtool{} is a research and auditing artifact, not a reference
encyclopedia. Every article is AI-generated and may be inaccurate,
fabricated, conflated, or outdated; the site says so prominently.
Even a 100\% factuality score covers only the sampled atomic claims
(${\sim}$10 per article), and most of the ${\sim}$1.3M pages are
unaudited: \emph{insufficient} does not mean false, and
\emph{supported} does not mean permanently true. Users should consult
authoritative sources in consequential domains.

The project is not intended for profiling individuals or supporting
decisions about them; public figures appear only as natural
encyclopedic subjects, and detected private information is excluded
from releases where possible.

The persona experiments are controlled interventions for studying
framing. Persona outputs do not represent the authors' views and may
reproduce ideological or cultural biases in training data; they
should be read as comparative experimental outputs, not endorsed
accounts.

We release prompts, articles, verdicts, and code under CC~BY~4.0;
redistributed content should retain attribution and be clearly marked
as AI-generated. Article images come from the openly licensed
\href{https://huggingface.co/datasets/Knowledge-aware-AI/100k-GenAI-Images-GPTKB}{100k-GenAI-Images-GPTKB}
collection.

% ═══════════════════════════════════════════════════════════════

\bibliography{references}

\end{document}